\documentclass{article}
\usepackage{spconf,amsmath,graphicx}
\usepackage{algorithm}
\usepackage{algpseudocode}
\usepackage{graphicx}
\usepackage{hyperref}
\usepackage{textcomp}
\usepackage{xcolor}
\usepackage{algorithm}
\usepackage{algpseudocode}
\usepackage{tabularx}
\usepackage{booktabs}
\usepackage{multirow}
\usepackage{amssymb}  % for \mathbb
\usepackage{amsmath}  % for \mathcal

\usepackage{enumitem}
\setlist{nosep, leftmargin=14pt}

\usepackage{mwe} % to get dummy images

\def\model{Insite}

\title{\textbf{INSITE}: Labelling Medical Images Using Submodular Functions and Semi-Supervised Data Programming\\}

\name{
  \begin{tabular}{c}
    Akshat Gautam$^{\star \dagger}$ \qquad Anurag Shandilya$^{\star \dagger}$ \qquad Akshit Srivastava$^{\dagger}$ \qquad Venkatapathy Subramanian$^{\dagger}$ \\
    Ganesh Ramakrishnan$^{\dagger}$ \qquad Kshitij Jadhav$^{\dagger}$\thanks{$\star$ equal contributors}
  \end{tabular}
}

\address{$^{\dagger}$ Indian Institute of Technology - Bombay, Mumbai, Maharashtra, India}
\begin{document}
% \name{
%   Akshat Gautam$^{\star \dagger}$ \qquad Anurag Shandilya$^{\star \dagger}$ \qquad Akshit Srivastava$^{\dagger}$ \qquad Venkatapathy Subramanian$^{\dagger}$ \\
%   \qquad Ganesh Ramakrishnan$^{\dagger}$ \qquad Kshitij Jadhav$^{\dagger}$ \thanks{$\star$ equal contributors}}

% \address{$^{\dagger}$ Indian Institute of Technology - Bombay, Mumbai, Maharashtra, India}

% \address{$^{\dagger}$ Indian Institute of Technology - Bombay, Mumbai, Maharashtra, India

% \name{Akshat Gautam$^{\star \dagger}$ \qquad Anurag Shandilya$^{\star \dagger}$ \qquad Akshit Shrivastava$^{\dagger}$ \qquad Venkatapathy Subramanian$^{\dagger}$  } 

% \name{\qquad Ganesh Ramakrishnan$^{\dagger}$ \qquad Kshitij Jadhav$^{\dagger}$}

% \author{\IEEEauthorblockN{1\textsuperscript{st} }
%     \IEEEauthorblockA{\textit{dept. name of organization (of Aff.)} \\
%     \textit{name of organization (of Aff.)}\\
%     City, Country \\
%     email address}

%\ninept
%
\maketitle
\begin {abstract}
The necessity of large amounts of labeled data to train deep models, especially in medical imaging creates an implementation bottleneck in resource-constrained settings. In \textbf{\model} (label\textbf{IN}g medical image\textbf{S} us\textbf{I}ng submodular func\textbf{T}ions and s\textbf{E}mi-supervised data programming) we apply informed subset selection to identify a small number of most representative or diverse images from a huge pool of unlabelled data subsequently annotated by a domain expert. The newly annotated images are then used as exemplars to develop several data programming-driven labeling functions. These labelling functions output a predicted-label and a similarity score when given an unlabelled image as an input. A consensus is brought amongst the outputs of these labeling functions by using a label aggregator function to assign the final predicted label to each unlabelled data point. We demonstrate that informed subset selection followed by semi-supervised data programming methods using these images as exemplars perform better than other state-of-the-art semi-supervised methods. Further, for the first time we demonstrate that this can be achieved through a small set of images used as exemplars . 

\end{abstract}

\section{Introduction}
\label{sec:intro}

Healthcare industry generates a significant amount of data, particularly in the field of medical imaging due to advancements in data acquisition techniques such as MRI, CT scans, {\em etc} \cite{lecun2015deep}. Consequently, the medical domain is considered the next frontier for artificial intelligence and machine learning, where Deep Neural Networks (DNNs) are expected to play a significant role \cite{baraniuk2020science}. This expectation is not unfounded since DNNs learn non-linear relationships between input variables and outputs facilitated by automated feature extraction, reducing the reliance on manual feature engineering \cite{lecun2015deep}.
However, despite the impressive advancements by DNNs, several disadvantages such as the need for huge computational resources and a large amount of high-quality labeled data have impeded their widespread adoption, especially in resource-poor settings \cite{havaei2023efficient} \cite{sun2017revisiting}. The latter is several times a bottle-neck for low- and middle-income countries since data labeling, typically done by individuals, that produces the "ground truth" necessary for supervised machine learning problems and strongly linked to predictive accuracy, is resource intense and expensive~\cite{linjordet2019impact}. To get around this limitation researchers have come up with semisupervised learning where a large number of unlabeled images can be used to assist in training and thus further improve the performance, generalization, and robustness of the models. {However, semi-supervised learning is domain specific and has difficulty in handling noisy labels which can be overcome through utilizing data programming by leveraging weak supervision \cite{sam2023losses}}. 

\hspace{1cm}

\textbf{\large Our Contribution:} Recent advances in the semi-supervised learning approach involves a novel alternate paradigm driven by data programming (section \ref{section4.2_data-programming}) where one can combine human knowledge obtained through small labeled data points with weak supervision to generate labels for unlabelled data \cite{maheshwari2021learning}. This involves the creation of labeling functions or rules to assign labels driven by heuristics or expert domain knowledge. While labeling functions may not be accurate by themselves, they identify some signal in the data point of interest, and then the outcome of several such labeling functions are aggregated using a generative model to assign the final label \cite{abhishek2021spear, maheshwari2021learning}. In the present paper, \textbf{\model}, we demonstrate that data programming could facilitate the labeling of large unlabelled datasets, especially in the medical domain since medical expert-driven labeling rapidly becomes expensive. Further, for the first time, we demonstrate that this can be done by using small expert annotated data points as exemplars (section \ref{Section4.3-Our_approach}) to be compared with unlabelled data points to develop these nuanced labeling functions.  

\section{Related Work}
\label{sec:format}
Scarcity of large labeled datasets presents a persistent challenge in medical domain, prompting the exploration of alternative techniques beyond traditional supervised deep learning approaches. Recent studies have ventured into semi-supervised learning to address this issue. %For instance, Peikari et al \cite{peikari2018cluster} proposed a semi-supervised approach for breast cancer digital pathology image analysis, which identifies areas of high and low density in a multidimensional feature space to place decision boundaries. 
%Liu et al  \cite{liu2022medical} introduced the PLAB-GAN framework for medical image classification, which involves clustering unlabelled samples to labeled image cluster centers, generating pseudo-labels based on CNN features, and training a classifier. Gu {\em et. al.} \cite{gu2020semi} addressed the limitations of Semi-Supervised Random Forest by modifying its training procedure with graph-embedded entropy, improving information gain calculations, and incorporating graph Laplacian regularization. Another approach, Anti-Curriculum Pseudo-Labelling (ACPL)\cite{liu2022acpl}, selects informative unlabelled images using cross-distribution sample informativeness and combines model and K-nearest neighbor (KNN) classification for pseudo-labelling. 
A teacher-student model with similarity learning for breast cancer segmentation, introduced by Cheng et al\cite{cheng2020self}, operates under a semi-supervised framework with minimal noisy annotations. Shi et al.\cite{shi2020loss} proposed a graph-based temporal ensembling model (GTE) inspired by Temporal Ensembling which encourages consistent predictions under perturbations, exploiting the semantic information of unlabeled data enhancing model robustness to noisy labels. 
Zhou et al. \cite{zhou2020deep} proposed a consistency-based approach based on new Mean-teacher (MT) framework based on template-guided perturbation- sensitive sample mining. This framework consists of a teacher network and a student network. The teacher network is an integrated prediction network from K-times randomly augmented data.
Srinidhi et al. \cite{srinidhi2022self} also proposed a semi-supervise preprocessing-based framework that combines self-supervised learning with semi-supervised learning. They first proposed the resolution sequence prediction (RSP) self-supervised auxiliary task to pre-train the model through unlabeled data, and then they performed fine-tuning of the model on the labeled data

%SimCLR\cite{chen2020simple}, a contrastive learning framework, aims to maximize the similarity between augmented views of the same image and minimize the similarity between different images in a minibatch . It incorporates resolution sequence prediction (RSP) as a self-supervised auxiliary task for pre-training on unlabeled data and subsequently fine-tunes the model on labeled data. The trained model is then used for further semi-supervised training based on the teacher-student consistency framework. These innovative approaches in the medical domain demonstrate the evolving landscape of machine learning paradigms, addressing the challenge of limited labeled data to improve the accuracy and robustness of medical image analysis and classification.

\section{Methods}
\label{sec:Methods}
\begin{figure}[ht]
  \centering
  \includegraphics[width=\columnwidth]{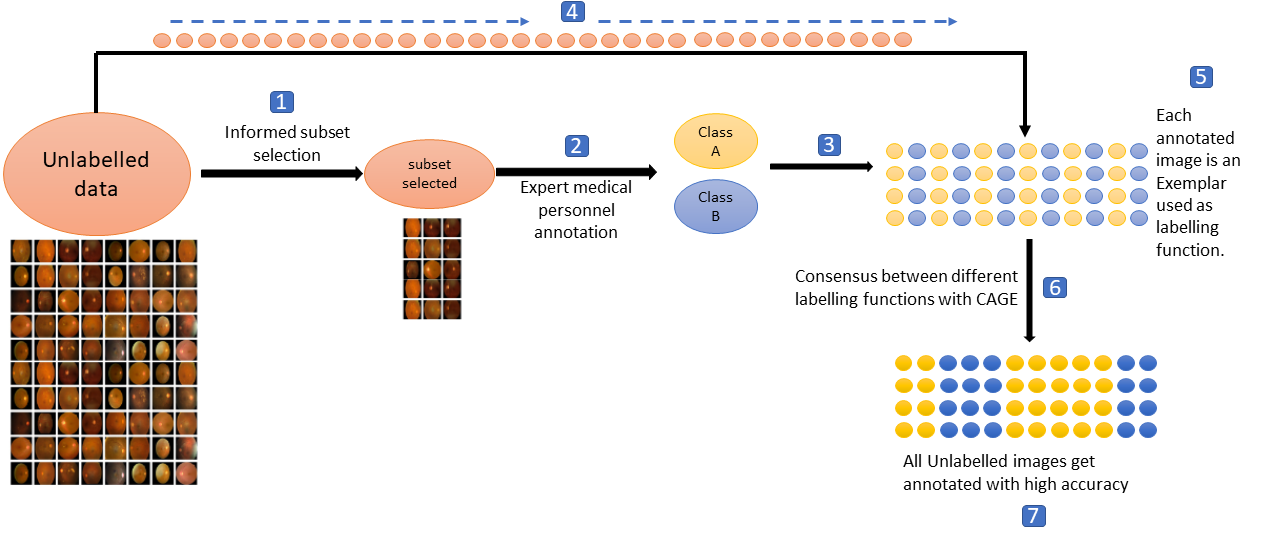} % Replace 'example-image-a' with your actual image file
  \caption{\footnotesize Submodular Function for subset selection, exemplars as labeling functions and CAGE \cite{chatterjee2019data} for label aggregation}
  \label{fig:cage}
\end{figure}

We start with a large amount of unlabelled data. Given the resource constraints, one can label only a small subset of images and instead of randomly selecting images to be labeled, we use submodular functions to identify a small number of most representative or diverse images which are then annotated by medical experts. These newly labeled images are then used as exemplars to develop several data programming-driven labeling functions each of which outputs a label and a similarity score when given an unlabelled image as an input. A consensus is brought about by using a label aggregator function to assign the final label to each unlabelled data point. Fig \ref{fig:cage} shows the schematic for our proposed approach.

\subsection{Submodular Functions} 
\label{Section4.1-submodular_functions}
Consider a ground-set $\mathcal{V}$ consisting of $n$ data points, denoted as $\mathcal{V} = \{1, 2, 3, \ldots, n\}$. We examine a set function $f: 2^{\mathcal{V}} \rightarrow \mathbb{R}$ that operates on subsets of $\mathcal{V}$ and outputs a real value. To qualify as submodular, the function $f$ adheres to the principle of diminishing marginal returns~\cite{fujishige2005submodular}. Specifically, for any two subsets $\mathcal{A} \subseteq \mathcal{B} \subseteq \mathcal{V}$ and an element $j \notin \mathcal{B}$, the function satisfies $f(j | \mathcal{A}) \geq f(j | \mathcal{B})$.

Different submodular functions capture various properties and objectives. For instance, the facility location function 

\[
f(\mathcal{A})=\sum\limits_{i \in \mathcal{V}} \max\limits_{j \in \mathcal{A}} S_{ij}
\]
aims to identify a representative subset by summing up the maximum pairwise similarity values $S_{ij}$. On the other hand, the log determinant function 

\[
f(\mathcal{A})=\log\det(S)
\]

seeks to construct a diverse subset, where $S$ is a matrix containing pairwise similarity values $S_{ij}$~\cite{iyer2015submodular}. A greedy algorithm was used which starts with an empty set and iteratively adds items that have the highest marginal gain to the set until the budget is exhausted.

\subsection{Data Programming:CAGE}
\label{section4.2_data-programming}
Let $\mathcal{X}$ denote the space of input instances and $\mathcal{Y} = \{0,1,\ldots,\linebreak K\}$ denote the space of labels, and let $P(\mathcal{X}, \mathcal{Y})$ denote their joint distribution. The goal is to learn a model to associate a label $y$ with an example $x \in \mathcal{X}$. Let the sample of $m$ unlabelled instances be $\{x_1,x_2,\ldots,x_m\}$. Instead of the true $y$’s, we are provided with a set of $n$ labeling functions (LFs) $\lambda_1, \lambda_2, \ldots, \lambda_{n}$, such that each LF $\lambda_j$ can be either discrete or continuous. Each LF $\lambda_j$ is associated with a class $k_j$ and, on an instance $x_i$, outputs a discrete label $\tau_{ij} = k_j$. If $\lambda_j$ is continuous, it also outputs a score $s_{ij} \in (0, 1)$. To learn the true label $y$, we use a generative model known as CAGE~\cite{chatterjee2019data} to aggregate heuristic labels by creating a consensus amongst LFs outputs. The generative model imposes a joint distribution between the true label $y$ and the values $\tau_{ij}, s_{ij}$ returned by any LF $\lambda_j$ on any data sample $x_i$ drawn from the hidden distribution $P(\mathcal{X} , \mathcal{Y})$. \cite{chatterjee2019data} define the hidden distribution as:

\begin{equation}\footnotesize
P(y,\tau_{ij}, s_{ij})= \frac{1}{Z_\theta}
\prod_{j=0}^{k-1} \psi_\theta(\tau_{ij},y)(\psi_\pi(\tau_{ij},s_{ij},y))^{cont(\lambda_j)}
\end{equation}

\noindent where cont($\lambda_j$) is 1 when $\lambda_j$ is a continuous LF and 0 otherwise, and $\theta$, $\pi$ denote the parameters used in defining the potentials $\psi_\theta$, $\psi_\pi$, coupling the discrete and continuous variables respectively. All LFs are independent of each other since each model is trained separately.   

For designing the potentials, \cite{chatterjee2019data} suggest that:

\begin{equation}\footnotesize
  \psi_\theta(\tau_{ij},y) =
    \begin{cases}
      \exp(\theta_{jy}) & \text{if $\tau_{ij} \neq 0$}\\
      1 & \text{otherwise}
    \end{cases}       
\end{equation}

\begin{equation}\footnotesize
  \psi_\pi(\tau_{ij},s_{ij},y) =
    \begin{cases}
      \text{Beta}(s_{ij} ; \alpha_a, \beta_a)  & \text{if $k_{j} = y$ \& $\tau_{ij} \neq 0$,} \\
      \text{Beta}(s_{ij} ; \alpha_d, \beta_d)  & \text{if $k_{j} \neq y$ \& $\tau_{ij} \neq 0$,} \\
      1 & \text{otherwise}
    \end{cases}       
\end{equation}

\noindent where $\text{Beta}$ density is expressed in terms of $\alpha,\beta>0$ as $P(s|\alpha, \beta) \propto s^{\alpha - 1}(1-s)^{\beta - 1}$. \cite{chatterjee2019data} replace $\alpha,\beta$ with alternative parameters such that 
$\alpha_a = q^c_j\pi_{jy}$ and $\beta_a = (1 - q^c_j)\pi_{jy}$ are parameters of the agreement distribution, and
$\alpha_d = (1 - q^c_j)\pi_{jy}$ and $\beta_d = q^c_j\pi_{jy}$  are parameters of the disagreement distribution. With these potentials, the normalizer $Z_\theta$ of our joint distribution (Eqn 1) can be calculated as:

\begin{equation}\footnotesize
    Z_\theta = \sum_{y \in \mathcal{Y}} \prod_j (1+ \exp(\theta_{jy}))
\end{equation}

For training the aggregation model, we maximize likelihood on the observed $\tau_i$ and $s_i$ values of the training sample $D = \{x_1, \dots, x_m\}$ after marginalizing out the true $y$. The training objective can be expressed as:

\begin{equation}\footnotesize
\max_{\theta, \pi} LL(\theta, \pi \vert D) 
\end{equation}
\begin{equation}\footnotesize
\begin{aligned}
LL(\theta, \pi|D) &= \sum_{i=1}^m \log \sum_{y \in Y} P_{\theta, \pi}(\tau_i, s_i, y) \\
&= \sum_{i=1}^m \log \sum_{y \in Y} \sum_{j=1}^n \left( \psi_j(\tau_{ij}, y) \cdot \right. \\
&\quad \left. \psi_j(s_{ij}, \tau_{ij}, y)\right) ^{\text{cont}(\lambda_j)}  - m \log Z_{\theta}
\end{aligned}
\end{equation}

 %\begin{figure*}[ht]
  %\resizebox{\linewidth}{6.0 cm}%{\includegraphics{Flowchart.png}}
 % \caption{Submodular Function for subset %selection, exemplars as labeling functions and %CAGE \cite{chatterjee2019data} for label %aggregation}
 % \label{fig:cage}
 %\end{figure*}

\subsection{Our Approach (Algorithm \ref{algo:insite})} 
\label{Section4.3-Our_approach}Let $X$ represent the set of input images, and $Y={0,1,\ldots,K}$ represent the set of labels. With a limited budget $b$, we employ informed subset selection to identify a subset of size $b$ that best represents the dataset or identifies the most diverse images. This subset is carefully labeled by a medical domain expert and denoted as $L={x_1, x_2, \ldots, x_b}$. The remaining unlabeled images form the set $U={u_1, u_2, \ldots, u_{|X|-b}}$, such that $X = U \cup L$. Pre-trained RESNET was utilized to extract features of each image in $X$. To facilitate the labeling process, $b$ labeling functions (LFs) are created, denoted as $\lambda_{1}, \lambda_{2}, \ldots, \lambda_{b}$. Here, each expert annotated image $L={x_1, x_2, \ldots, x_b}$ is used as an exemplar. Each LF $\lambda_i$ is associated with a corresponding image $x_i \in L$. Given an unlabelled input image $u_j$, the LF $\lambda_i$ provides two outputs: the label $l_i$ associated with $x_i$, and the correlation score $s_i \in (-1,1)$ between the features extracted by the RESNET model for images $x_i$ and $u_j$. For each image $u_j \in U$, we obtain $b$ labels ($l_{u_{j}1}, l_{u_{j}2}, \ldots, l_{u_{j}b}$) and similarity scores ($s_{u_{j}1}, s_{u_{j}2}, \ldots, s_{u_{j}b}$) from the corresponding LFs. To infer the true label $l_{u_{j}}$ for an image $u_j$, we employ a generative model that aggregates labels by achieving a consensus among the outputs of the LFs to improve the accuracy of label assignment for the unlabeled images in $U$.

\label{algorithms}
\setlength{\textfloatsep}{0pt}
\begin{algorithm}\small
\caption{\texttt{\hspace{-7pt}\textbf{: Submodular function based subset selection, labelling using CAGE}}}\small
\label{algo:insite}
\begin{algorithmic} [1]
\Require Initial Image Set $X$, Budget: $B$, Labelled Set:$L=\emptyset$
\State Define the submodular function for eg. Facility Location $f(L) = \sum_{i \in X} \sum_{j \in L} d_{ij}$, where $d_{ij}$ is the distance between facilities $i$ and $j$

\State $L \gets \operatorname{argmax}_{L \subseteq X, |L| \le B} \sum_{x \in L} f(x)$
\State $U \gets X \setminus L$
\State Domain Expert labels images in $L$ to create $T$, such that $t_i$ is the label of $x_i \in L$
\State Extract features $\text{vec}(x)$ for each image $x \in L \cup U$

\State For each $x_u \in U$, calculate ${d(x_u, x_l)}_{x_l \in L}$, the set of distances to labeled images
\State $LabellingFunctions \gets$ Create LFs $\lambda_i$ for each $x_i \in L$
\State Using CAGE generative model, aggregate labels and similarities
\State Obtain final labels for images in $U$
\State \Return Final labels for images in $U$
\end{algorithmic}
\end{algorithm}
\section{Datasets and Results}
\label{sec:Datasets}
In this section, we discuss the results of our framework on several medical imaging datasets. The results demonstrate that \textbf{\model{}} is able to achieve high accuracy even with 1\% labeled data. We hypothesize that this is due to the selection of the most representative or diverse images driven by nuanced informed sub-set selection. These images form a seed for developing exemplar-driven labeling functions building on a data programming paradigm in a semi-supervised setting. We compare the accuracy of our method against state-of-the-art semi-supervised methods, which also deal with small labeled datasets. These methods also require labelled images to start with, which are given both randomly and using informed subset selection. We convert benchmark datasets to a binary classification problem to develop a screening technique for distinguishing between normal and diseased conditions. 

\subsection{Dataset: APTOS 2019 Diabetic Retinopathy}
The APTOS 2019 Blindness Detection Challenge~\cite{karthick2019aptos} consists of retinal fundoscopy images which are categorized into five classes: normal (class 0:1,805), mild (class 1:370), moderate (class 2:999), severe (class 3:193), and proliferative (class 4:295). For our experiment, we combined images from classes 1, 2, 3, and 4 to create a single diseased
class (class 1: 1,857 images) of Diabetic Retinopathy.

\begin{table}[ht]\footnotesize
\centering

\begin{tabular}{l|cccc}
\toprule
Budget & 10 & 20 & 30 & 40 \\
\midrule
\textbf{FacilityLocation+Cage (our method)} & \textbf{0.84} & \textbf{\underline{0.91}} & \textbf{\underline{0.93}} & \textbf{0.92} \\
FacilityLocation+Cheng et al & 0.73 & 0.85 & 0.83 & 0.86 \\
FacilityLocation+Shi et al & 0.75 & 0.79 & 0.83 & 0.84 \\
FacilityLocation+Zhou et al & 0.71 & 0.80 & 0.82 & 0.85 \\
FacilityLocation+Srinidhi et al & 0.72 & 0.86 & 0.86 & 0.88 \\
\midrule
\textbf{LogDeterminant+Cage (our method)} & \textbf{\underline{0.89}} & \textbf{0.89} & 0.87 & 0.89 \\
LogDeterminant+Cheng et al & 0.74 & 0.79 & 0.83 & \textbf{0.90} \\
LogDeterminant+Shi et al & 0.69 & 0.81 & 0.84 & 0.83 \\
LogDeterminant+Zhou et al & 0.75 & 0.74 & \textbf{0.88} & 0.88 \\
LogDeterminant+Srinidhi et al & 0.75 & 0.83 & 0.84 & 0.86 \\
\midrule
Random+Cage & 0.64 & 0.67 & 0.75 & 0.8 \\

Random+Cheng et al & 0.85 & 0.90 & 0.90 & \textbf{\underline{0.94}} \\
Random+Shi et al & \textbf{0.88} & \textbf{\underline{0.91}} & \textbf{0.91} & 0.93 \\
Random+Zhou et al & 0.84 & 0.87 & 0.89 & 0.90 \\
Random+Srinidhi et al & 0.85 & 0.88 & 0.91 & 0.93 \\
\bottomrule
\end{tabular}
\caption{\footnotesize Accuracy on APTOS dataset for different budgets}

\label{Table:1}
\end{table}

The results on APTOS Dataset (Table \ref{Table:1}) reveal that both FacilityLocation and LogDeterminant functions combined with CAGE (\model) demonstrate robust performance, with the former outperforming all baseline measures at budget levels of 20 and 30. At a budget of 10, the LogDeterminant function appears to yield superior outcomes. Upon increasing the budget to 40, the semi-supervised approach proposed by Cheng et al. achieves the highest accuracy, although FacilityLocation (\textbf{Insite}) also performs fairly well.

\subsection{Chest X-Ray Images}
Daniel S. {\em et .al.}~\cite{kermany2018identifying} curated a large and diverse dataset of 5,856 Chest X-ray images for the diagnosis of pediatric pneumonia. The dataset is divided into 2 classes - Normal and Pneumonia. 
Table \ref{Table:2} compares the accuracies of \textbf{\model{}} and baselines on the Chest X-Ray dataset on different budgets for subset selection
\begin{table}[ht]\footnotesize
\centering

\begin{tabular}{l|cccc}
\toprule
Budget & 10 & 20 & 30 & 40 \\
\midrule
\textbf{FacilityLocation+Cage (our method)} & \textbf{\underline{0.87}} & \textbf{\underline{0.86}} & 0.86 & 0.8 \\
FacilityLocation+Cheng et al & 0.80 & 0.82 & \textbf{\underline{0.88}} & \textbf{0.90} \\
FacilityLocation+Shi et al & 0.73 & 0.82 & 0.85 & 0.87 \\
FacilityLocation+Zhou et al & 0.79 & 0.83 & 0.85 & 0.85 \\
FacilityLocation+Srinidhi et al & 0.77 & 0.83 & 0.85 & 0.89 \\
\midrule
\textbf{LogDeterminant+Cage (our method)} & 0.76 & \textbf{\underline{0.86}} & \textbf{0.86} & 0.87 \\
LogDeterminant+Cheng et al & 0.76 & 0.80 & 0.84 & \textbf{\underline{0.91}} \\
LogDeterminant+Shi et al & 0.74 & 0.85 & 0.86 & 0.90 \\
LogDeterminant+Zhou et al & 0.69 & 0.78 & 0.83 & 0.83 \\
LogDeterminant+Srinidhi et al & \textbf{0.77} & 0.84 & 0.86 & 0.89 \\
\midrule
Random+Cage & 0.68 & 0.68 & 0.82 & 0.77 \\

Random+Cheng et al & 0.72 & 0.82 & 0.83 & 0.85 \\
Random+Shi et al & 0.75 & 0.81 & 0.83 & 0.83 \\
Random+Zhou et al & \textbf{0.77} & 0.80 & 0.81 & 0.82 \\
Random+Srinidhi et al & 0.75 & \textbf{0.83} & \textbf{0.84} & \textbf{0.86} \\
\bottomrule
\end{tabular}
\caption{\footnotesize Accuracy on CHEST dataset for different budgets}
\label{Table:2}
\end{table}

\noindent For the Chest X-Ray dataset, the application of the FacilityLocation submodular function for the seed set indetification followed by CAGE driven labelling surpasses all baseline metrics at budget allocations of 10, 20. The semi-supervised method by Cheng {\em et. al.} gives better accuracy for a budget of 30 and 40 images selected for labelling as the seed set.

\subsection{Tuberculosis TB X-Ray Images:}
In their study, Rahman et al. \cite{rahman2020reliable}constructed a repository of chest X-ray images, which includes both Tuberculosis (TB) positive and normal cases. 

\begin{table}[ht]\footnotesize
\centering

\begin{tabular}{l|cccc}
\toprule
Budget & 10 & 20 & 30 & 40 \\
\midrule
\textbf{FacilityLocation+Cage (our method)} & \textbf{\underline{0.89}} & \textbf{\underline{0.88}} & \textbf{0.88} & \textbf{0.90} \\
FacilityLocation+Cheng et al & 0.72 & 0.82 & 0.83 & 0.85 \\
FacilityLocation+Shi et al & 0.75 & 0.82 & 0.85 & 0.85 \\
FacilityLocation+Zhou et al & 0.79 & 0.80 & 0.83 & 0.87 \\
FacilityLocation+Srinidhi et al & 0.75 & 0.81 & 0.87 & 0.86 \\
\midrule
\textbf{LogDeterminant+Cage (our method)} & 0.57 & 0.59 & 0.64 & 0.76 \\
LogDeterminant+Cheng et al & 0.77 & 0.84 & 0.89 &\textbf{ 0.90} \\
LogDeterminant+Shi et al & 0.69 & 0.78 & 0.83 & 0.84 \\
LogDeterminant+Zhou et al & \textbf{0.80} & \textbf{0.83} & \textbf{0.87} & 0.88 \\
LogDeterminant+Srinidhi et al & 0.75 & 0.79 & 0.81 & 0.85 \\
\midrule
Random+Cage & 0.58 & 0.63 & 0.88 & 0.91 \\
Random+Cheng et al & 0.71 & 0.72 & 0.87 & 0.89 \\
Random+Shi et al & 0.57 & 0.59 & \textbf{\underline{0.90}} & \textbf{\underline{0.91}} \\
Random+Zhou et al & 0.65 & 0.69 & 0.81 & 0.86 \\
Random+Srinidhi et al & \textbf{0.75} &\textbf{0.83} & 0.84 & 0.86 \\
\bottomrule
\end{tabular}

\hspace{1cm}
\caption{\footnotesize Accuracy on TB dataset for different budgets}
\label{Table:3}
\end{table}
\noindent For the given dataset, the FacilityLocation for selecting the seed set of images for expert annotation combined with CAGE continues to outperform baseline approaches with budgets of 10, 20. At the increased budget level of 30,40, the technique introduced by Shi et al. attains the highest accuracy (Table \ref{Table:3}). 

Our methods (informed subset selection followed for CAGE(data programming + label aggregation)) outshine several semi-supervised benchmarks, particularly at lower budgets of 10, and 20. As the budget increases, alternative approaches begin to show enhanced performance. This may be because neural networks can leverage a larger budget more effectively, learning from the increased data. Nonetheless, our method is specifically designed for scenarios with very few labeled medical images. Importantly, informed subset selection for seed set annotation followed by CAGE was overall better than informed or random subset selection combined with other semi-supervised methods (indicated by underscored results in Tables \ref{Table:1}, \ref{Table:2}, \ref{Table:3}) especially in scenarios of low budgets.  Additionally, our technique does not rely on neural network training, except for the utilization of pre-trained ResNet weights for feature extraction. 

\section{Conclusion}
Our proposed semi-supervised method represents a significant leap in medical image classification, particularly when confronted with the challenge of limited labelled images. INSITE demonstrates superior accuracy, particularly in scenarios characterized by stringent budget constraints for image labeling. This work holds immense promise, particularly in resource-scarce settings where obtaining a substantial volume of labelled data remains a formidable challenge. However, it is important to acknowledge the current limitation of our method, which primarily focuses on binary datasets. Future endeavors could enhance the versatility of our approach by extending its applicability to multi-class datasets.

\section{Compliance with ethical standards}
There are no conflict of interest. Datasets are publicily accessible, they are all licensed and properly cited.
\bibliographystyle{IEEEbib}
\bibliography{refs}

\end{document}